# AI for Porosity and Permeability Prediction from Geologic Core X-Ray Micro-Tomography

**Zangir Iklassov**  
MBZUAI

**Dmitrii Medvedev**  
MBZUAI

**Otabek Nazarov**  
MBZUAI

**Shakhboz Razzokov**  
Gubkin University of O&G In Tashkent

## Abstract

Geologic cores are rock samples that are extracted from deep under the ground during the well drilling process. They are used for petroleum reservoirs' performance characterization. Traditionally, physical studies of cores are carried out by the means of manual time-consuming experiments. With the development of deep learning, scientists actively started working on developing machine-learning-based approaches to identify physical properties without any manual experiments. Several previous works used machine learning to determine the porosity and permeability of the rocks, but either method was inaccurate or computationally expensive. We are proposing to use self-supervised pretraining of the very small CNN-transformer-based model to predict the physical properties of the rocks with high accuracy in a time-efficient manner. We show that this technique prevents overfitting even for extremely small datasets.
Github: https://github.com/Shahbozjon/porosity-and-permeability-prediction

## 1 Introduction

Rock cores are cylindrical specimens of the formations down the well-bore. They are drilled out of the reservoir and extracted to the surface. After that scientists traditionally carry out experiments in the laboratories to find out rock's physical properties like porosity and permeability. *Porosity* is the ratio of the void space to solid space inside the core, while *permeability* is the ability of the rock to pass fluid through itself. These two properties are crucial for all the stages of petroleum reservoir deployment. Based on them, engineers make the estimates of the total petroleum fluid present in the reservoir, calculate the potential productivity rate of the reservoir, decide the most optimal grid for wells construction, etc. Hence, cores' physical analysis is an important step in the reservoir development pipeline.

Recently, the digital study of the rock cores' physical properties became prevalent due to technological developments. Scientists study rock properties by employing X-ray micro-tomography and numerical simulation methods. The general calculation process includes three steps. In the first step, X-ray micro-tomography is carried out to get the 3D images of the rock samples. During the second step, the segmentation is performed to help with the classification of all the kinds of rocks present in the core sample along with the classification of porous and non-porous formations. In the last step, different mathematical and physical computations are carried out to determine the desired properties.

To the best of our knowledge, the last step is most often carried out using deterministic differential equations [9]. There have been several attempts to apply machine learning approaches for predicting porosity and permeability. Some of the works describe a mix of traditional computational methods along with deep learning models [11]. These techniques are very slow, thus the only way to speed them up is by giving up on accuracy. Another set of research relies on 3D CNNs. Models perform well on synthetic datasets when cores are generated using Gaussian distribution, but the accuracy does not generalize to real-life cores [6]. When the 3D CNN models are used on the real datasets, they tend to show severe overfitting due to the small dataset size [7].

To tackle the issues listed above we developed a new compact CNN-transformer-based model (1168 neurons). Due to the small size of the available dataset, we first pretrained the model in a self-supervised manner. We chose a self-supervised task of predicting the masked regions of input cubes because it would make the model learn the inner pore structures of the rock. Then, we use pretrained weights for predicting the porosity and permeability of the given rock. Ablation study shows that our technique prevents overfitting and produces a model with good results on an extremely small dataset.

## 2 Literature Review

Traditionally scientists use algorithms of flow simulation in the porous media when they want to rely on computational powers rather than manual experiments. The most common methods are pore network models (PNMs), semi-analytical Laplace solvers (SASs), and the lattice Boltzmann method (LBM) [9]. Each of the methods provides different accuracy and it comes with the trade-off of computational time it requires.

With the development of deep learning (DL) researchers started testing out the applicability of deep learning algorithms on this task. Traditional algorithms are quite slow, while DL has better computational inference speed.

One of the first studies looked at the relationship between permeability and other rock properties like porosity, pore geometry, connectivity, etc [4]. Some correlation models based on multi-variate regression have been proposed but then they required substantial preliminary information about the rock's physical properties.

One of the first uses of neural networks was in conjunction with the traditional LBM method [11]. LBM provided permeability measurements of 2D cross-sections. Those measurements along with a few other geometric properties went into a single hidden layer as input and permeability was calculated. The model gave $R^2$ value of 0.996. This accuracy was achieved due to LBM already doing most of the calculations. The advantage of this method came from LBM's faster calculation of 2D cross-sections compared to 3D cross-sections.

Tian et. al. generated a set of synthetic rock core samples [8]. They used 16 different geometric measures of the rocks as inputs. Neural network with 9 hidden layers achieved $R^2$ of 0.999, but the dataset they used was synthetic and their homogeneous samples were coming from the uniform distribution with a narrow permeability range of 0.3 to 1.2 D [permeability measurement unit].

In another study 1,100 rock sub-samples were used (400 samples with geometries of 152x152x175 and 759 with geometries of 100x100x160) [6]. They tested different machine learning algorithms including linear regression, gradient boosting, random forest, multi-layer perceptron (MLP), and convolutional neural networks (CNN). As input to the models, they used tortuosity, porosity, and permeability calculations of PNM (fastest but the most inaccurate deterministic method). MLP and CNN gave the best results of approximately $R^2 = 0.91$ each.

Instead of relying on the traditional geometric characterization of rocks, some scientists relied on using convolutional neural networks to let models extract rock geometry characterization itself [7]. One of the works used 5-layer CNN to predict porosity values. They tried feeding into CNN both binary (already segmented into solid and pore spaces) and gray-scale 2D image slices. Binary images gave $R^2 > 0.90$, while gray-scale ones gave $R^2 > 0.75$.

One of the first works to predict permeability using 2D CNNs used Voronoi Mesh-based geometries to pre-process data and get better porous representations [10]. And only using 2 convolutional layers along with 3 dense layers, they predicted permeability values with an accuracy of $R^2 = 0.92$ on the training set, but only $R^2 = 0.20$ on the testing set.

In another work authors used 2D CNN as pseudo-3D CNN using the gray-scale images and feeding neighboring rock sample slices as separate channels [5]. It was a 3 layer CNN with 2 dense layers at the end. They generated synthetic data of 17,700 rocks with 256x256x256 dimensions. They got good results of $R^2 = 0.9385$. However, they haven't compared the performance of other methods like simpler multi-layer perceptron on their synthetic dataset. Moreover, it is hard to confidently state the success of their model because most models performed well on synthetic datasets in general.

There were several studies that directly used 3D CNNs. The difference between them is the small architectural changes and the dataset they utilized. The one that used synthetic dataset got $R^2 = 0.91$ on the test set [3], while when the same model used real dataset got $R^2 = 0.91$ on the train set but $R^2 = 0.69$ on the test set [2]. Hence, in the case of a synthetic dataset, the model is probably not 'blind' and encounters test set samples in the train set. Since simple 3D CNN model is not generalizable as it performs poorly on the test set, the third model used more complex architecture (3D ResNet and ResNext), and their real rocks dataset permeability values had a small range between 0.0001 D and 1.8 D. They got $R^2 = 0.87$ and $R^2 = 0.86$ on the train and test sets respectively [1].

## 3 Datasets

The datasets will be taken from `digitalrocksportal.org`. The name of the dataset is *11 Sandstones: raw, filtered, and segmented data*. It contains 11 X-ray rock samples with 1000x1000x1000 dimensions each presented both in grayscale and binary formats. In the binary case, the data is already segmented into porous and non-porous types.

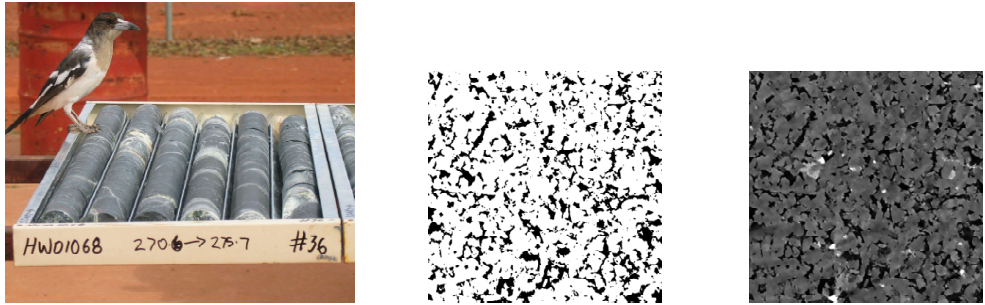

Figure 1: Left - core examples; center - binary X-ray core slice; right - grayscale X-ray core slice

## 4 Methodology

The goal of our model is to produce physical predictions in an accurate and fast manner. Accounting for rock's time-series type pore spatial distribution is very important because permeability is directly related to how well pores are connected along with the rock sample. Therefore, we designed our own architecture that takes into account the time-series type of relationship within the core (see Figure 1). Initially, using CNN we extract spatial features of the core. Then we pass CNN feature maps to the transformer's multi-head attention layer. If CNN initially extracts important small-scale features of rock slices, the transformer accounts for a general porous structure within the rock.

### 4.1 Self-Supervised Pretraining

Due to the small size of the available labeled data, we decided to first pretrain the model in a self-supervised manner, and then use it for porosity and permeability predictions. Additionally, to increase the amount of data, we adopted the following technique. The input data - a 10x10x10 size sample, sub-cube of the core - is randomly drawn from the core's 3D scan. Before proceeding, random parts of the slices of the sub-cube are masked, and after that, we apply 3x3x10 convolution.

Then, convolutional maps of 10x10x10 size are sent to multi-head self-attention layers of the transformer model, which account for the general contextual information of the sub-cube. This helps to model interrelations between the sub-cubes inside the core sample.

A 10x10x10 output of multi-head attention is passed to a fully-connected layer, which produces the restoration of the initial input core. RMSE loss is used for the model's optimization.

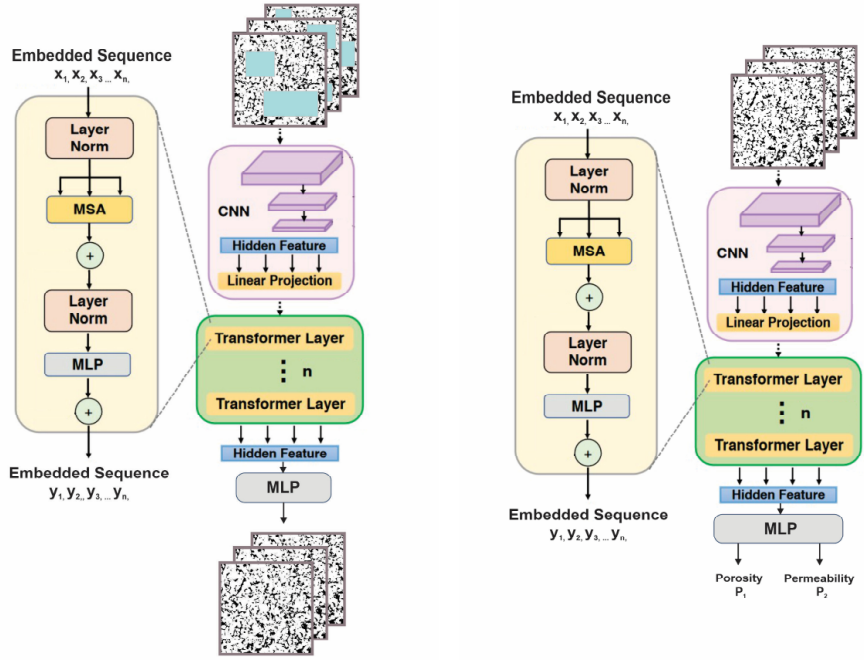

Figure 2: Models' architectures (left - SSL model; right - supervised model)

### 4.2 Permeability and Porosity Prediction

After the pretraining stage, we can use the same model for the main task of permeability and porosity prediction. The only architectural change would be replacing the MLP head of pretrained model with another MLP head that would predict two values - permeability and porosity (see Figure 1, right side). The supervised model's architecture is initialized with pretrained model's weights, except the MLP head which is randomly initialized. Also, 10x10x10 subcubes would be used as input but this time without any mask. RMSE loss is used again for the model's optimization.

## 5 Experiments and Results

### 5.1 Self-Supervised Pretraining

We built our own architecture from scratch, using grid search and 5-fold cross-validation to find a model with the lowest RMSE value. For each hyperparameter, we created an interval from which we randomly drew values and trained networks with corresponding architecture comparing them to the same test dataset. Values for the number of neurons, number of layers, activation functions, and batch size were randomly selected. The following parameter combinations were tested: learning rate 0.0001 to 1, with powers of 10 increments, 1 to 5 layers for each type of the layer (convolutional, multi-head attention, fully connected), 2 to 64 feature maps in each convolutional layer, ReLU, sigmoid, and tanh activation functions, 16 - 512 batch sizes with powers of 2 increments. RMSE was used as a main metric and $R^2$ as a secondary metric for model evaluation.

The following configuration showed the best performance: ReLU activation function, batch size 128 and learning rate 0.001; two convolutional layers (each having 10 feature maps and 3x3 kernel size); one multi-head self-attention layer with 10 attention heads; 2 fully-connected layers with 64 and 125 neurons respectively, the latter makes final prediction of 10x10x10 core.

The initial dataset was randomly split into 10x10x10 size sub-cube samples, each consisting of 1000 numerical values. A total of 1 million data points, or sub-cubes, jointly accounted for equally dispensed training and testing sets. Each sub-cube had 20% of its data masked and processed through the model to output the sub-cube with the same dimensions as input.

Table 1: Performance of the model with different hyperparameters

| Learning rate | Epochs | Activation Function | Total number of neurons | Number of layers |
| Batch size | Training loss | Training $R^2$ | Testing loss | Testing $R^2$ |
|---|---|---|---|---|
| 0.001 | 30 | ReLU | 1168 | 5 |
| 128 | 0.02100 | 0.84907 | **0.02094** | **0.84995** |
| 0.01 | 25 | Tanh | 1176 | 6 |
| 64 | 0.02267 | 0.84121 | 0.02184 | 0.84281 |
| 0.1 | 35 | ReLU | 1184 | 8 |
| 256 | 0.02144 | 0.8474 | 0.02674 | 0.82677 |
| 0.01 | 20 | Sigmoid | 1568 | 10 |
| 512 | 0.02491 | 0.80894 | 0.02791 | 0.83274 |
| 0.001 | 30 | ReLU | 1368 | 7 |
| 32 | 0.02561 | 0.8249 | 0.02876 | 0.80782 |

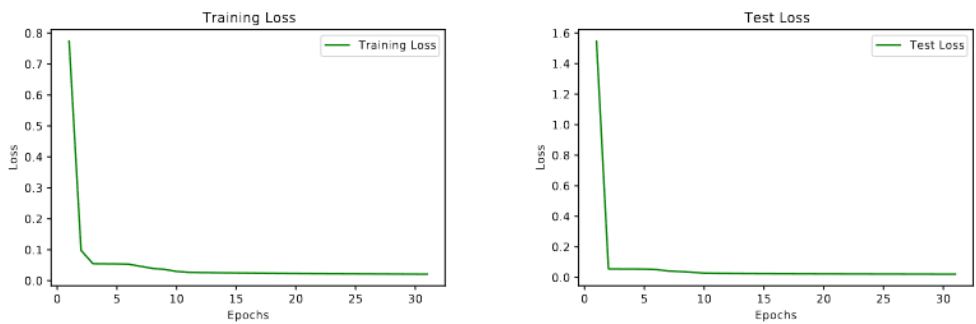

Figure 3: Self-supervised losses (left - training loss; right - testing loss)

The plots in Figure 3 demonstrate the abrupt decrease of both training and testing losses after the first epoch with the following asymptotic tendency to values close to zero. Any notable difference between both results is missing, which means there was no overfitting during a training phase. Specifically, the training and testing errors are 0.021003 and 0.020940 respectively.

It is worth mentioning, that we have 1 million data points for each of two sets - training and testing - while our network accounts for 1168 neurons. In other words, the dataset is by 3 orders larger than the number of learnable parameters; therefore, there is a very low chance of overfitting. And finally, the mean of the test masked target values is 0.7 with a standard deviation is 0.4, while the RMSE values are much smaller compared to these statistics. Since RMSE is much smaller than standard deviation we can conclude that model results are statistically significant with $R^2$ of 0.84995.

## 5.2 Porosity and Permeability Prediction

Table 2: Performance of supervised model with random and pretrained initializations

| Initialization | Training Loss | Testing Loss | Training $R^2$ | Testing $R^2$ |
|---|---|---|---|---|
| Random | 10.6 | 12.9 | 0.7029 | 0.6743 |
| Pretrained | 10.6 | 10.6 | 0.7031 | 0.7024 |

We made new sampling using the entire dataset of size 11x1000x1000x1000 (eleven core scans) for supervised training purposes. The first six cores were used as part of a training set giving one million sub-cubes, while the rest were used as part of the testing set. Thus, we had an almost equal train and test split. However, as output, we still had only 11 unique values.

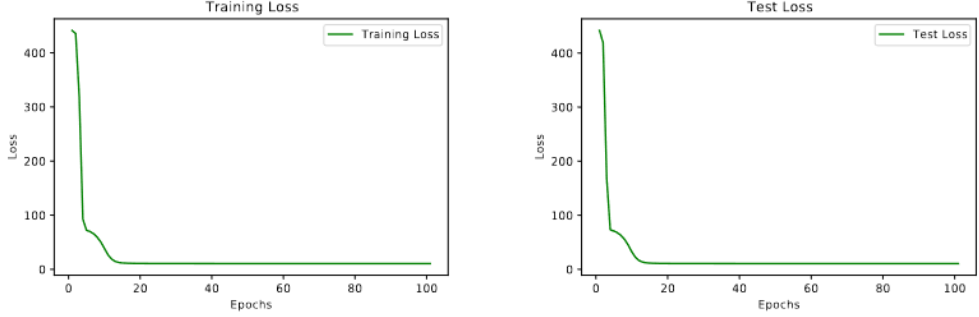

Figure 4: Porosity and Permeability losses (left - training loss; right - testing loss)

Table 3: Results of different methods from literature review on the task of permeability prediction

| Model | Dataset size | Train $R^2$ | Test $R^2$ | Inference time |
|---|---|---|---|---|
| LBM | - | 0.98+ | 0.98+ | 1 - 4 days |
| SAS | - | 0.85 - 0.95 | 0.85 - 0.95 | 12 - 24 hrs |
| PNM | - | 0.80 - 0.90 | 0.80 - 0.90 | 1 - 4 hrs |
| 3D CNN | 3158 | 0.91 | 0.69 | 1 s |
| 3D ResNet | - | 0.87 | 0.86 | 1 s |
| Ours (random) | 11 | 0.70 | 0.67 | 1 s |
| Ours (pretrain) | 11 | 0.70 | 0.70 | 1 s |

The test data for porosity had the following statistics: mean = 20.89; max = 26.54; min = 14.77; std = 3.2625. Test data for permeability had: mean = 168.1034; max = 386; min = 10; std = 126.1015.

We employed the same hyperparameter settings as with the self-supervised case, except the learning rate. After checking various learning rates ranging from 10e-6 to 10e-2, we have found that the learning rate of 10e-5 showed the best result on the test set.

The results showed a test error of 10.6 and $R^2$ of 0.7032 (see Table 2). These are average metrics of both output variables. The results on the training dataset are similar: 10.606 and 0.7031 respectively. Based on the error and metric values, and training curves (see Figure 4) we can conclude that there is no overfitting. The error value is greater than the standard deviation of porosity and less than the standard deviation of permeability.

Additionally to check the effectiveness of our self-supervised approach we have trained the model with exact same hyperparameters, but with random initialization. As a result, we have observed that testing loss and $R^2$ are more than training ones (see Table 2). Thus, our self-supervised pretraining approach solves the issue of overfitting.

And the model we have developed has advantages when compared to other existing methods. First of all, it is much faster than deterministic methods (see the first group of rows in Table 3). Also, other deep learning methods have been trained on thousands of rock samples (see the second group of rows in Table 3), yet they had severe overfitting and have similar $R^2$ as our model which has been trained only on 11 samples. Note, 3D ResNet in the Table 3didn't report the dataset size, but we hypothesize that it must have had a very large dataset to train a big model like 3D ResNet without overfitting. Additionally, the data range for permeability in the case of the 3D ResNet model varied between 0-100 mD, while ours varied between 0-380 mD. Thus, the 3D ResNet model worked with a dataset that had smaller variance and would inherently tend to show higher $R^2$ values.

## 6 Discussion and Conclusion

In our work, we have developed a new AI technique, which can benefit the oil and gas industry. Our deep learning model allows inferring the two most fundamental properties of the formation - porosity and permeability - without the need to run expensive and time-consuming experiments in the laboratory. Unlike the classical approach which is based on hard-coded equations and requires up to several days to derive the desired properties of the down-hole formation, our methodology allows inferring porosity and permeability in a matter of seconds. Furthermore, our methodology did not show any signs of overfitting in the experiments we have carried out.

Increasing the variance of the training dataset can be of benefit for further validating our model. In our work, we used the dataset consisting only of 11 core samples, which is on the order of $10^3$ smaller than what was used in the previous works of other deep learning researchers. Even though we obtained results on par with previous deep learning works, it is necessary to apply models on exactly the same datasets of a much larger size to obtain a statistically significant comparison.